%% file: root.tex
\documentclass[letterpaper, 10 pt, conference]{ieeeconf}  

\IEEEoverridecommandlockouts                              

\usepackage{graphics} 
\usepackage{epsfig} 
\usepackage{mathptmx} 
\usepackage{times} 
\usepackage{amsmath} 
\usepackage{amssymb}  
\usepackage{algorithmic}
\usepackage{algorithm}
\usepackage{mathtools}

\usepackage{siunitx}
\usepackage{booktabs} 
\usepackage{graphicx}

\usepackage{subcaption}
\newtheorem{prop}{Proposition}
\makeatletter
\let\NAT@parse\undefined
\makeatother
\usepackage[bookmarks=true]{hyperref}

\title{Constrained Reinforcement Learning with Smoothed Log Barrier Function}

\author{Baohe Zhang\authorrefmark{2},~Yuan Zhang\authorrefmark{2},~
Lilli Frison\authorrefmark{4},~Thomas Brox\authorrefmark{2}\authorrefmark{3},~Joschka Bödecker\authorrefmark{2}\authorrefmark{3}
\thanks{
\authorrefmark{2}: Department of Computer Science, University of Freiburg, Germany \authorrefmark{3}: BrainLinks-BrainTools, University of Freiburg, Germany
\authorrefmark{4}: Fraunhofer Institute for Solar Energy Systems, Germany.
        Contact: {\tt\footnotesize zhangb@cs.uni-freiburg.de}}
}

\begin{document}

\maketitle

\begin{abstract}
\input{0_abstract}
\end{abstract}

\section{INTRODUCTION}
\input{1_introduction}

\section{RELATED WORK}
\input{2_related}
\section{BACKGROUND}
\input{3_backgrounds}

\section{APPROACH}
\input{4_approach}
\section{RESULTS}
\input{5_results}
\section{CONCLUSION}
\input{7_conclusion}





\clearpage
\bibliographystyle{IEEEtran}
\bibliography{references}

\end{document}

%% file: 0_abstract.tex
Reinforcement Learning (RL) has been widely applied to many control tasks and substantially improved the performances compared to conventional control methods in many domains where the reward function is well defined.
 However, for many real-world problems, it is often more convenient to formulate optimization problems in terms of rewards and constraints simultaneously.
Optimizing such constrained problems via reward shaping can be difficult as it requires tedious manual tuning of reward functions with several interacting terms.
Recent formulations which include constraints mostly require a pre-training phase, which often needs human expertise to collect data or assumes having a sub-optimal policy readily available.
We propose a new constrained RL method called CSAC-LB ($\textbf{C}$onstrained $\textbf{S}$oft $\textbf{A}$ctor-$\textbf{C}$ritic with $\textbf{L}$og $\textbf{B}$arrier Function), which achieves competitive performance without any pre-training by applying a linear smoothed log barrier function to an additional safety critic. It implements an adaptive penalty for policy learning and alleviates the numerical issues that are known to complicate the application of the log barrier function method.
As a result, we show that with CSAC-LB, we achieve state-of-the-art performance on several constrained control tasks with different levels of difficulty and evaluate our methods in a locomotion task on a real quadruped robot platform.

%% file: 1_introduction.tex
In traditional Reinforcement Learning~\cite{DBLP:books/lib/SuttonB98} formulations,  rewards are usually the only metric of performance.
However, in many real-world applications, the performance of an algorithm is not necessarily measured in terms of a single objective.
For example, considering an autonomous driving task, the agent needs to drive as fast as possible but still needs to respect the traffic rules and avoid collisions.
Designing a reward function which considers all individual cases is often difficult. Tuning the weights of different reward components might also lead to a sub-optimal solution.
In the context of deep reinforcement learning, where neural networks are used as function approximators, it is often not feasible to apply conventional constrained optimization methods due to the large number of parameters to be optimized and the limited extrapolation capabilities of neural networks~\cite{158898}.

\begin{figure}[t]
    \centering
    \includegraphics[width=\columnwidth]{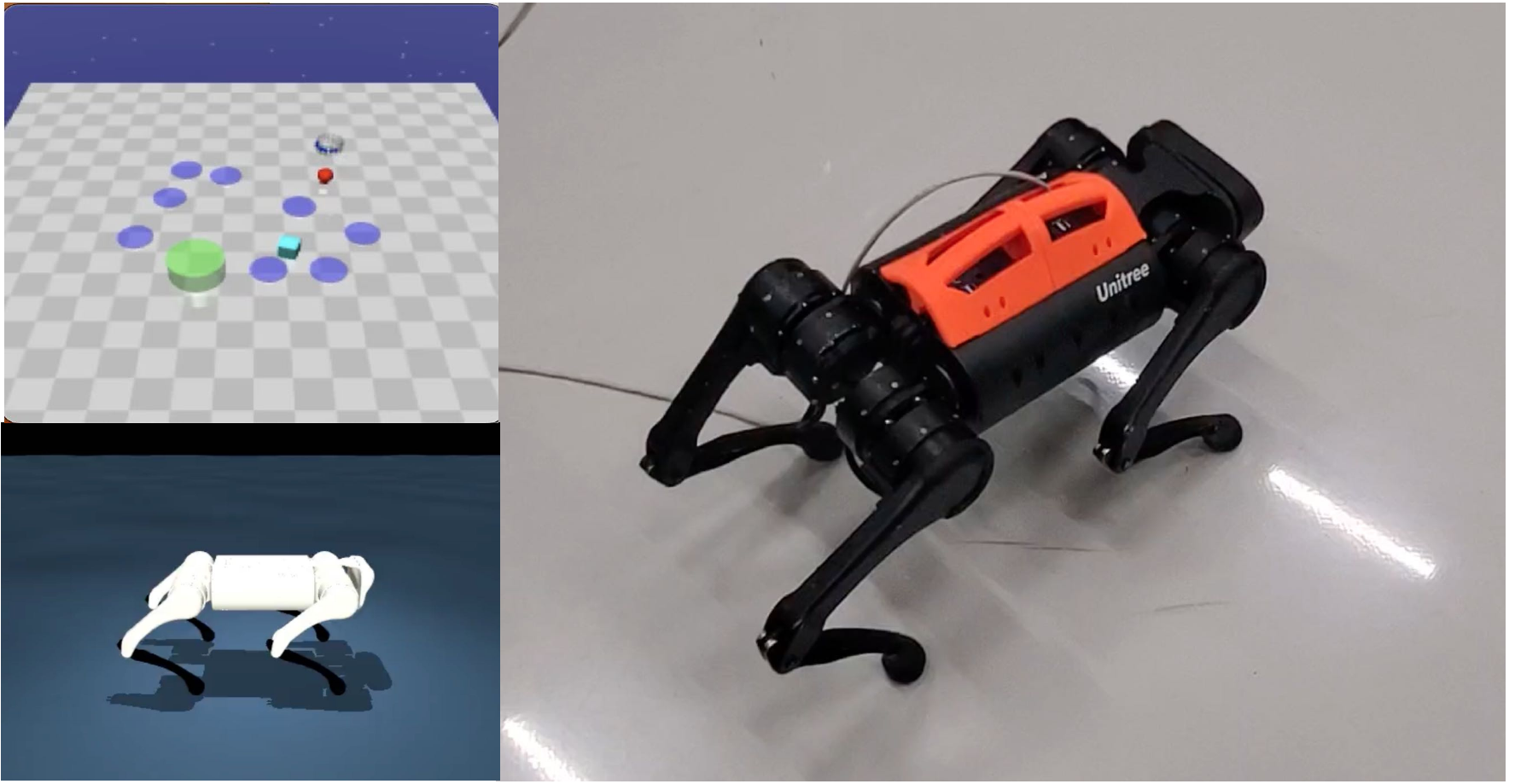}
    \caption{\textbf{Upper Left}: SafetyGym\cite{ray2019benchmarking} PointGoal1-v0 task. The red dot is the controlled agent. The agent is required to reach the goal area (green) and avoid going through blue pillars that are randomly generated. 
    \textbf{Lower Left}: Unitree A1~\cite{unitree2018unitree} in Mujoco\cite{todorov2012mujoco} Simulator with randomized terrain.
    \textbf{Right}: Unitree A1 with a protective shield from \cite{Wu22CoRL_DayDreamer}.}
    \label{fig:environments}
\vspace{-10pt}
\end{figure}

If no prior knowledge is available, constraint violation is unavoidable as the RL agent needs to visit the unsafe states of the environment to gradually learn a safe, but possibly conservative policy. This is often inherently contradictory to the primary objective of obtaining more rewards.
A concomitant challenge is to predict the boundary of the region of safe states (i.e. of the feasible set) with a neural network for a given state-action pair as neural networks typically cannot extrapolate well to unseen data~\cite{158898}. To predict the safe margin, the RL agent must first see some samples from the unfeasible set. However, dramatically violating the constraints is also generally undesirable, as this may cause severe damage to the system being optimized. This means that an RL agent should try to avoid large constraint violations and learn efficiently from as few constraint violations as possible.

Considering these challenges, existing approaches often require assumptions, such as the availability of a known sub-optimal policy, or prior knowledge about the dynamics or the cost function. This limits the generalization of the algorithm to new tasks and requires human expertise.
Especially in tasks where the optimal solution lies on the safe margin, to improve data efficiency and allow for an accurate prediction of the safe margin, the agent needs to explore sufficiently along the safe margin.

Motivated by the above, we propose CSAC-LB, which applies a linear smoothed log barrier function~\cite{logbarrierExtension} to the Soft Actor-Critic (SAC) algorithm~\cite{DBLP:conf/icml/HaarnojaZAL18} augmented by a safety critic~\cite{DBLP:journals/corr/safety_critic}. It acts as a general-purpose method that can handle different types of cost functions without any additional information or training. Being an off-policy algorithm, it can leverage data collected previously for efficient learning. The goal of this algorithm is to effectively explore the safe margin of a given problem during training and to learn a well-performing policy for future deployment in real-world tasks. This distinguishes it from other works that aim for zero constraint violations at training time. Our contributions are the following:
\begin{itemize}
    \item We propose CSAC-LB, an off-policy model-free method which can handle the numerical issues commonly associated with the log barrier method. The algorithm is easy to implement and can serve as a general-purpose method for constrained RL problems. A theoretical performance boundary of the algorithm is also provided.
    \item We conduct a variety of experiments on comparing CSAC-LB with other SOTA baselines on different high-dimensional control tasks. Our algorithm gives the best overall performance in all tasks.
    \item We demonstrate the robustness of CSAC-LB via zero-shot sim2real policy transfer in a locomotion task. Our algorithm is the only one among all tested baselines that achieved successful transfer to the robot.
\end{itemize}

%% file: 2_related.tex
Many prior works have been proposed for addressing the safety of Reinforcement Learning using various approaches.
The reviews in~\cite{DBLP:journals/jmlr/GarciaF15,DBLP:journals/ml/Dulac-ArnoldLML21,DBLP:conf/l4dc/MaLLZC22} may be referred to for a comprehensive overview.

One direction which has been explored for solving constrained RL problems is safe policy search.
For instance, techniques from nonlinear programming are integrated into a policy gradient approach~\cite{4354030}.
In the context of lifelong RL~\cite{DBLP:conf/icml/Bou-AmmarTE15} builds up a framework motivated by theoretical analysis for safe policy search via gradient projection.
Constrained Policy Optimization~(CPO)~\cite{DBLP:conf/icml/AchiamHTA17} is the first general-purpose approach that solves constrained reinforcement learning using a trust-region method and is able to provide certain theoretical guarantees.
In \cite{polymenakos2019safe}, the authors extend safe policy search by augmenting it with a Gaussian Process model for risk estimation.
A recent line of research~\cite{DBLP:conf/cdc/LiFST21} uses the Lagrange-multiplier method to propose a new objective, whereas~\cite{DBLP:conf/nips/ChowTMP15,DBLP:journals/jmlr/ChowGJP17} use Conditional Value-at-risk (CVaR) as the metric and optimize the Lagrangian function with gradient descent.

In \cite{DBLP:conf/corl/HaXTLT20}, the authors extend SAC~\cite{DBLP:conf/icml/HaarnojaZAL18} with a cost function and optimize the constrained problem by introducing a Lagrange-multiplier.
However, this method suffers in terms of training robustness when constraint violations occur only rarely during training on the task.
With this framework, they are able to learn multiple tasks with constraints on a real-world robot.
Using a safety critic, \cite{DBLP:journals/corr/safety_critic} learn to predict the cumulative constraint violations.
Further improvements are presented in \cite{DBLP:conf/aaai/YangSTS21} where a distributional safety critic is combined it with the CVaR metric.
By changing the percentile of this CVaR metric, the sensitivity to risk can also be changed accordingly.
The work presented in \cite{DBLP:conf/ijcai/YingZ0Y0022} also uses CVaR as the metric, but in conjunction with the on-policy method PPO~\cite{DBLP:journals/corr/SchulmanWDRK17} as the RL algorithm.
Lyapunov functions are also widely applied to constrained RL problems~\cite{DBLP:conf/nips/ChowNDG18, DBLP:journals/corr/abs-1901-10031}.
By projecting policy parameters onto feasible solutions from linearized Lyapunov constraints during the policy update, these methods can be applied to any policy gradient method, such as DDPG~\cite{DBLP:journals/corr/LillicrapHPHETS15} or PPO.

Another line of work is close to Model Predictive Control (MPC)~\cite{mpc}.
Model-based approaches, such as~\cite{DBLP:conf/nips/BerkenkampTS017, DBLP:conf/aaai/ChengOMB19}, use a dynamics model to certify the safety of the system. These approaches usually assume that the dynamics model is calibrated and the constraints are known. This, however, limits the general applicability of the algorithms. A framework for learning barrier certificates~\cite{8796030} and the policy iteratively, achieving zero training-time constraint violations in an empirical analysis, is presented in~\cite{DBLP:conf/nips/LuoM21}. The work in~\cite{DBLP:conf/corl/Pereira0ET20} combines stochastic barrier functions with safe trajectory optimization and is able to recover the optimal policy under certain conditions.
In~\cite{DBLP:conf/iclr/AsUC022}, the authors improve data efficiency in constrained RL by using a Bayesian World model.

Close to our work is the application of interior-point methods to Proximal Policy Optimization~(PPO)~\cite{DBLP:journals/corr/abs-1812-06502,DBLP:conf/aaai/LiuDL20}.
However, these methods are on-policy methods and have numerical stability issues due to the log barrier function.
To the best of our knowledge, we introduce the first numerically stable off-policy algorithm using the log barrier function for the constrained RL setting and address the training robustness of constrained RL.

%% file: 3_backgrounds.tex
\subsection{Constrained Markov Decision Processes}
A Markov Decision Process~(MDP) is defined as a tuple $(\mathit{S}, \mathit{A}, \mathit{R}, \mathit{\gamma}, \mathit{P})$, where $\mathit{S}$ and $\mathit{A}$ are a set of states and a set of actions, respectively.
$\mathit{R} : \mathit{S} \times \mathit{A} \rightarrow \mathbb{R}$ is the reward function, $\mathit{\gamma}$ is a discount factor and $\mathit{P}$ is the probabilistic state transition model.
To construct a Constrained Markov Decision Process (CMDP) \cite{altman1995constrained}, we further augment an MDP by adding a cost function $C : \mathit{S} \times \mathit{A} \rightarrow \mathbb{R}^{\mathit{K}}$ which implements a set of constraints and returns a $\mathit{K}$-dimensional vector of costs given a state-action pair.
We define $\mathit{J}_\mathit{C}(\pi)$ as the expected cumulative discounted cost of a given policy $\pi$ and $\mathit{D}$ is a $\mathit{K}$-dimensional vector of cost limits.
The feasible set of policies for a CMDP can then be written as
\begin{align}
\Pi_{\mathit{C}} = \{\pi \in \Pi : \mathit{J}_\mathit{C} (\pi) -  \mathit{D} \leq 0\}
\end{align}

We denote $\mathit{J}(\pi)$ as the expected cumulative return to be maximized.
Then, the optimization problem of learning an optimal policy can be written as
\begin{align}
\pi^{*} = \arg \max_{\pi \in \Pi_{\mathit{C}}} \mathit{J}(\pi)
\end{align}
which maximizes the return and respects all constraints.

\subsection{SAC-Lagrangian}
\label{sec:sac-lagrangian}

\paragraph{Lagrange multiplier method} This method~\cite{BERTSEKAS1982179} has been widely used in constrained optimization problems. We define a constrained optimization problem as:
\begin{align}
\label{eq:ori_cons_problem}
\max_{x} f(x) ~~~~ \textrm{s.t.} ~ g(x) \leq 0
\end{align}

By introducing an adaptive Lagrange multiplier $\beta$ in the optimization, Eq.~\ref{eq:ori_cons_problem} can be written as an unconstrained max-min Lagrangian function:
\begin{align}
\label{eq:lag_eq}
\max_{x}\min_{\beta \geq 0} \mathit{L} (x, \beta) \doteq f(x) - \beta g(x)
\end{align}

\paragraph{SAC-Lagrangian (SAC-Lag)}
SAC-Lag is a modified version of Soft Actor-critic~\cite{DBLP:conf/icml/HaarnojaZAL18} and was first designed in the context of learning locomotion for quadruped robots~\cite{DBLP:conf/corl/HaXTLT20}. To prevent possible damage to the robot, constraints are added at each step to limit the pose of the robot. To formulate the constrained problem, we denote $d_{t}$ as the cost limit for constraint violations at step $t$ and the objective to maximize can be written as:
\begin{align}
\label{eq:rl_cons_prob}
& \sum^{T}_{t=0} \mathop{\mathbb{E}}_{\mathit{a}_t \sim \pi (\mathit{s}_t)} \biggr[ \gamma^{t} \mathit{R}(\mathit{s}_t, \mathit{a}_t) + \alpha \mathit{H}(\pi(\cdot | \mathit{s}_t))\biggr ] \\
& \textrm{s.t.} ~  \mathop{\mathbb{E}}_{\mathit{a}_t \sim \pi (\mathit{s}_t)} \biggr[C(\mathit{a}_{t}, \mathit{s_{t}}) - d_{t} \biggr] \leq 0, ~ \forall t \nonumber
\end{align}
where $\mathit{R}$ is the reward function and $\mathit{H}(\pi(\cdot | \mathit{s}_t))$ is the entropy term.
\cite{ray2019benchmarking} extend this method to cumulative constraints and, similar to~\cite{DBLP:journals/corr/safety_critic}, they apply two critic networks to learn the cumulative discounted reward and the exceeding cost of constraint violations, respectively.
We denote the reward Q-network with parameters $\theta_{r}$ as $\mathit{Q}_{\theta_{R}}$ and the cost Q-network with parameters $\theta_{c}$ as $\mathit{Q}_{\theta_{c}}$.
This formulation also allows multitask training without the need to retrain the cost network, i.e., the cost Q-network does not need retraining when transferring to a new task with the same constraints. Now with the Lagrange-multiplier method, denoting $d$ as the cumulative cost limit, the unconstrained optimization problem can be formulated as:
\begin{align}
\label{dual_sac}
& \max_{\pi} \min_{\beta \geq 0}\mathit{L}(\pi, \beta) \doteq f(\pi) - \beta g(\pi)
\end{align}
\begin{align*}
& \textrm{where} ~ & & f(\pi) = \sum^{T}_{t=0} \mathop{\mathbb{E}}_{\mathit{a}_t \sim \pi (\mathit{s}_t)} \biggr[ \gamma^{t} \mathit{R}(\mathit{s}_t, \mathit{a}_t) + \alpha \mathit{H}(\pi(\cdot | \mathit{s}_t))\biggr] \nonumber \\
& \textrm{and} ~ & & g(\pi) = \sum^{T}_{t=0} \mathop{\mathbb{E}}_{\mathit{a}_t \sim \pi (\mathit{s}_t)} \biggr[\gamma^{t} (C(\mathit{a}_{t}, \mathit{s_{t}}) - d )\biggr] \nonumber
\end{align*}

Since neural networks are used to approximate the Q-value, there is no closed-form solution for the Eq.~\ref{dual_sac}. Dual Gradient descent~\cite{boyd_vandenberghe_2004} is then used to iteratively update the policy and the Lagrange multiplier. Denote the parameters of the actor network with $\phi$ and the replay buffer $\mathit{D}$.

To minimize Eq.~\ref{dual_sac}, Lagrange-multiplier $\beta$ is updated according to the following loss:
\begin{align}
\mathit{J}(\beta) = \mathop{\mathbb{E}}_{\mathit{s}_t \sim \mathit{D}, \mathit{a}_t \sim \pi_{\phi} (\mathit{s}_t)} \biggr[ \beta (d - \mathit{Q}_{\theta_{c}}(\mathit{s}_t, \mathit{a}_t))\biggr ]
\end{align}

When the cost Q-network output is beyond cost limit $d$, $\beta$ will increase to enhance the constraints and decrease when constraints are expected to be satisfied by the current policy.
Now, assuming a fixed value for $\beta$, the actor loss to be minimized from  Eq.~\ref{dual_sac} can be written as:
\begin{multline}
    \mathit{J}(\phi) = \mathop{\mathbb{E}}_{\mathit{s}_t \sim \mathit{D}, \mathit{a}_t \sim \pi_{\phi} (\mathit{s}_t)} \biggr[ \alpha \log \pi_{\phi}(\mathit{a}_t | \mathit{s}_t) - \\ \mathit{Q}_{\theta_{r}}(\mathit{s}_t, \mathit{a}_t) + \beta \mathit{Q}_{\theta_{c}}(\mathit{s}_t, \mathit{a}_t)\biggr ]
\end{multline}

%% file: 4_approach.tex
\subsection{Log Barrier Method}
Here, we introduce the log barrier method as the fundamental method of our approach to solve inequality-constrained problems as defined in Eq.~\ref{eq:ori_cons_problem}. In contrast to the usage of Lagrange multipliers, the log barrier method transforms a constrained problem into an unconstrained problem by augmenting the objective function with a so-called log barrier function $\psi (x) = - \frac{1}{\mu} \log (-x)$, where $\mu$ is the log barrier factor to change the accuracy of the approximation to an indicator function, which gives an infinite penalty when $x > 0$ and remains $0$ when $x \leq 0$. With the log barrier method, the unconstrained problem can be formulated as:
\begin{align}
\label{eq:log_barrier_method}
\max_{x} \mathit{L} (x) \doteq f(x) - \psi (g(x))
\end{align}
The benefit of the log barrier function is that it approximates the non-differentiable indicator function with a continuous differentiable function and still prevents the constraint term from actively contributing for values below $0$.
The log barrier function acts similarly to the Lagrange multiplier, which adaptively relaxes or enhances the constraints. But its penalty increases exponentially when constraints are strongly violated, thus leading to a better recovery from unsafe states during the optimization.


\begin{figure}[ht]
    \centering
    \includegraphics[width=\columnwidth]{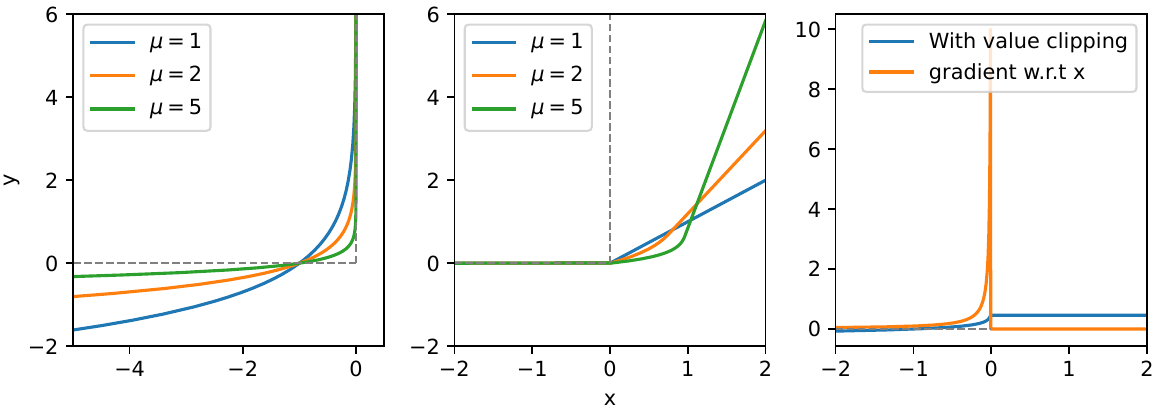}
    \caption{\textbf{Left}: Log barrier function with different $\mu$. The dashed line is the indicator function. When $x \rightarrow 0$, $\psi (x) \rightarrow \infty$, it leads to infinity penalty. When $\mu \rightarrow \infty$, the log barrier function $\psi (x)$ is close to a step function
    \textbf{Mid}: Linear Smoothed Log barrier function with different $\mu$. The dashed line is the indicator function. When $\mu \rightarrow \infty$, the linear smoothed log barrier function $\Tilde{\psi} (x)$ is close to a step function with infinity penalty when $x > 0$
    \textbf{Right}: Log barrier function (Blue curve) with Value Clipping. Due to the value clipping when $y=10$, the gradient vanishes.}
    \label{fig:ext_log_barrier}
    \vspace{-4pt}
\end{figure}

\subsection{Linear Smoothed Log Barrier Function}
The log barrier method can serve as a powerful tool to solve constrained problems with inequality constraints. However, it is known to suffer from numerical stability issues. When the feasible set is empty, the optimization becomes unstable as the logarithm function does not allow for $g(x) > 0$. This becomes more problematic when we use gradient descent to optimize the problem as it can not guarantee the constraints to be satisfied in each step.

In the context of Deep Reinforcement Learning, neural networks are also used to approximate the value functions and represent the continuous control policy. Due to the random initialization of the neural network, the actor network may initially not satisfy all constraints and thus represent an unsafe policy. One possible solution then is to clip the values or to apply an activation function such as ReLU~\cite{DBLP:conf/icml/NairH10} to the output layer. However, this would also stop the gradient flow when a constraint violation occurs and even cause complete ignorance of the constraint violation since the objective receives no penalty for running out of the feasible set.

To address the numerical stability issue and tackle the issues from directly clipping the output of the neural network,~\cite{logbarrierExtension} propose a linear smoothed log barrier function $\Tilde{\psi}(x)$, shown in Fig.~\ref{fig:ext_log_barrier}, defined as: 
\begin{align}
\label{eq:ext_log_barrier}
\Tilde{\psi}(x) = \begin{cases}
       - \frac{1}{\mu} \log (-x) & \text{if $x \leq -\frac{1}{\mu^{2}}$}\\
      \mu x - \frac{1}{\mu}\log (\frac{1}{\mu^{2}}) + \frac{1}{\mu} & \text{otherwise}\\
    \end{cases}       
\end{align}
Because $\Tilde{\psi}(x)$ is continuous and differentiable everywhere, we are not restricted only to the feasible set and can apply stochastic gradient descent~(SGD) to optimize the neural networks.

\paragraph{CSAC-LB}
With the help of the linear smoothed log barrier function, we are able to solve the constrained RL problem defined in Eq.~\ref{eq:rl_cons_prob} via SGD. We follow the setup of SAC-Lag to have two double-Q~\cite{DBLP:conf/aaai/HasseltGS16} critic networks $\mathit{Q}_{\theta^{1}_{r}}, \mathit{Q}_{\theta^{2}_{r}}$ and $\mathit{Q}_{\theta^{1}_{c}}, \mathit{Q}_{\theta^{2}_{c}}$ to learn the reward and the cost of constraint violation, respectively. For simplicity, we use the same parameter $\theta_r$ or $\theta_c$ to refer to the double-Q network of reward or cost. 
The linear smoothed log barrier function has a non-zero small gradient even if constraints are satisfied, which may lead to a sub-optimal solution.
To overcome this problem, we apply a ReLU activation at the input of $\psi(x)$, shift $+1$ along the positive axis, and denote this modified version as $\Tilde{\psi}^{*}(x)$, where $d$ is the cost limit and $\mu > 1$:
\begin{align}
\label{eq:ext_log_barrier_rl}
\Tilde{\psi}^{*}(x) = \psi(ReLU(x-d)-1)  
\end{align}
For simplicity, we define $\Tilde{\mathit{J}}$ as the new objective $\mathit{J}^{LB}$ after plugging in the log barrier function The objective to be minimized for the actor network becomes
\begin{multline}
\label{eq:csac_loss}
\Tilde{\mathit{J}}(\phi) = \mathop{\mathbb{E}}_{\mathit{s}_t \sim \mathit{D}, \mathit{a}_t \sim \pi_{\phi} (\mathit{s}_t)} \biggr[ \alpha \log \pi_{\phi}(\mathit{a}_t | \mathit{s}_t) \\ - \mathit{Q}_{\theta_{r}}(\mathit{s}_t, \mathit{a}_t) + \Tilde{\psi}^{*}(\mathit{Q}_{\theta_{c}}(\mathit{s}_t, \mathit{a}_t))\biggr ]
\end{multline}

We denote with $\theta_r'$ and $\theta_c'$ the parameters of the target networks for reward and cost, respectively. The critic loss to train reward and cost networks is the same as in standard SAC, which is:
\begin{align}
\label{eq:csac_critic_reward_loss}
\Tilde{\mathit{J}}(\theta_r) = \mathop{\mathbb{E}}_{(\mathit{s}_t, \mathit{a}_t) \sim \mathit{D}} \biggr[ (\mathit{Q}_{\theta_r}(\mathit{s}_t, \mathit{a}_t) - \hat{Q}_{\theta_r'}(\mathit{s}_t, \mathit{a}_t))^2 \biggr ] 
\end{align}
\begin{align}
\label{eq:csac_critic_cost_loss}
\Tilde{\mathit{J}}(\theta_c) = \mathop{\mathbb{E}}_{(\mathit{s}_t, \mathit{a}_t) \sim \mathit{D}} \biggr[ (\mathit{Q}_{\theta_c}(\mathit{s}_t, \mathit{a}_t) - \hat{Q}_{\theta_c'}(\mathit{s}_t, \mathit{a}_t))^2 \biggr ]
\end{align}
\begin{align*}
& \textrm{where} ~ & & \hat{Q}_{\theta_r'}(\mathit{s}_t, \mathit{a}_t) = r(\mathit{s}_t, \mathit{a}_t) + \gamma \mathit{Q}_{\theta_r'}(\mathit{s}_{t+1}, \mathit{a}_{t+1}) \nonumber \\
&  & & \hat{Q}_{\theta_c'}(\mathit{s}_t, \mathit{a}_t) = c(\mathit{s}_t, \mathit{a}_t)  + \gamma \mathit{Q}_{\theta_c'}(\mathit{s}_{t+1}, \mathit{a}_{t+1})   \nonumber\\
& \textrm{and} ~ & & \mathit{a}_{t+1}\sim \pi_{\theta_a}(s_{t+1}) \nonumber 
\end{align*}
The complete algorithm can be found in Algorithm~\ref{alg:csac-lb}.

Because of the exponential increase of constraint violations, CSAC-LB allows the agent policy to recover quickly when it leaves the safe margin and becomes unsafe.
Whereas, for algorithms such as SAC-Lag, due to the dual-optimization of the Langrangian-multiplier, the agent needs extra training steps to return to a safe policy and can cause severe constraint violation during this time.

\begin{algorithm}[ht]
    \caption{CSAC-LB}
    \label{alg:csac-lb}
    \textbf{Input}: initial the double-Q critic network parameter $\theta_{r}$, $\theta_{c}$ and actor network parameter $\phi$, log barrier factor $\mu$, discount factor $\gamma$, entropy term $\alpha$, learning rate $\lambda$, polyak update factor $\tau$ \\
    \textbf{Initialize}: Target networks, $\theta_{r}' \leftarrow \theta_{r}$, $\theta_{c}'\leftarrow \theta_{c}$ \\
    \textbf{Initialize}: Replay buffer $\mathit{B}$
    \begin{algorithmic}[1] 
    \FOR{$t \gets 1$ to $T$}
    \STATE Sample action $a_{t} \sim \pi_{\phi}(s_{t})$
    \STATE Observe reward $r_{t}$, cost $c_t$ and next state $s_{t+1}$
    \STATE Store transition $(s_t, a_t, r_t, c_t, s_{t+1})$ in the replay buffer $\mathit{B}$
        \FOR{each gradient step}
            \STATE Sample a batch of transitions from replay buffer $\mathit{B}$
            \STATE Update reward critic network $\theta_{r} \leftarrow \lambda \nabla_{\theta_{r}} \mathit{J}^{LB}(\theta_{r})$
            \STATE Update cost critic network ~~~~~ $\theta_{c} \leftarrow \lambda \nabla_{\theta_{c}} \mathit{J}^{LB}(\theta_{c})$
            \STATE Update actor network ~~~~~~~~~~~~$\phi ~~ \leftarrow \lambda \nabla_{\phi}\mathit{J}^{LB}(\phi)$
            \STATE Update target networks: \\
            ~~~~ $\theta_{r}' \leftarrow \tau \theta_{r} + (1-\tau)\theta_{r}'$ \\
            ~~~~ $\theta_{c}' \leftarrow \tau \theta_{c} + (1-\tau)\theta_{c}' $
        \ENDFOR
    \ENDFOR \\
    \textbf{Output}: Optimized parameter $\theta_{r}$, $\theta_{c}$ and policy $\pi_{\phi}$
    \end{algorithmic}
\end{algorithm}

\paragraph{Performance Guarantee Bound}
We analyze the performance guarantee bound of the optimal value after applying the linear smoothed log barrier function.  For simplicity, we set the constraint limit~$d$ to 0.
The proof is omitted due to the space limit. As a reference, a similar derivation is achieved in~\cite{DBLP:journals/corr/abs-1812-06502}. 

\begin{prop}
The maximum gap between the optimal value of the constrained problem (Eq.~\ref{eq:ori_cons_problem}) and the optimal value obtained by minimizing the unconstrained objective defined in ~Eq.~\ref{eq:csac_loss} is bounded by 0, where $\mu > 1$, $m$ is the number of constraints and $\mu$ is the log barrier factor of the log barrier function, if the problem is feasible.
\end{prop}

%% file: 5_results.tex
\subsection{Baselines}
We use the following three baselines to compare with our algorithm:~(a) \textbf{SAC}~\cite{DBLP:conf/icml/HaarnojaZAL18} with reward shaping: we add a terminal reward of -30 when the constraint is violated, which is tuned by~\cite{DBLP:conf/nips/LuoM21}.~(b) \textbf{SAC-Lagrangian} \cite{ray2019benchmarking}: One of the state-of-the-art algorithms in Constrained RL as introduced in Sec.~\ref{sec:sac-lagrangian}.~(c) \textbf{WCSAC-0.5}~\cite{DBLP:conf/aaai/YangSTS21}: It extends SAC-Lag with a safety measurement, which is the level of conditional Value-at-Risk (CVaR) from the distribution.

Algorithms are re-implemented and trained with the same hyperparameters as summarized in Tab.~\ref{tab:common_hp}.

    \begin{table}[ht]
    	\begin{center}
    	\begin{small}
    		\begin{sc}
    		  \scalebox{0.8}{
			\begin{tabular}{lcll}
				\toprule
				\textbf{Hyperparameters} & \textbf{Value}\\
				\midrule
			batch size & 256\\
                Network Architecture & [256,256] \\
			discount factor $\gamma$ & 0.99 \\
			random steps & 100 \\
			learning rate & 1e-4 \\
			actor update frequency & 1 \\
                critic update frequency & 10 \\
			polyak update factor & 0.005 \\
			init temperature & 1.0 \\
                weights of cost in SAC & 30 \\
                log barrier factor in CSAC-LB & 2 \\
                safety level of WCSAC & 0.5 \\

				\bottomrule
			\end{tabular}
			}
    			\end{sc}
    		\end{small}
    	\end{center}
           \caption{The hyperparameters}
    	\label{tab:common_hp}
    \end{table}


\subsection{Simulation Experiments}
\paragraph{Environment Setup}
We use two simulated tasks, depicted in Fig.~\ref{fig:environments}, to evaluate the generalization of our algorithm CSAC-LB and the other baselines.
The first task is based on PointGoal1-v0 implemented in Safety Gym~\cite{ray2019benchmarking}.
The agent receives a binary cost indicating collision with the obstacles at every step.
This task has a 60-dimensional observation space and a 2-dimensional action space which moves the agent up/down and left/right, respectively.

We further evaluate our algorithm on a quadruped locomotion task to prove its performance and stability.
Inspired by~\cite{fuminimizing}, the task is transformed into a safe RL problem where the quadruped needs to minimize its energy consumption (reward) while maintaining its speed between $80\%$ and $120\%$ of the desired speed ($\nu_{d}$). 
Specifically, we use the Unitree A1 robot~\cite{unitree2018unitree} and the Mujoco simulator~\cite{todorov2012mujoco} as the experimental platform. The observation space and action space follow the previous work~\cite{smith2022walk} with 78 and 12 dimensions respectively. The observation space contains the IMU, velocity, and joint position. The action space is the desired position of each joint. The reward is given as the negative energy consumption, which is calculated by the joint torques of all motors. Regarding the speed constraint, we adopt $0.375m/s$ and $0.9m/s$. Domain randomization~\cite{peng2018simtoreal} is added to help bridge the gap between simulation and real robots.
We denote $\tau_{i}$ and $q_{i}$ as the torque and joint velocity of $i$-th joint on the robot.
The reward function and cost of violating the constraints are shown in Eq.~\ref{eq:reward} and Eq.~\ref{eq:cost}.
\begin{align}
reward = - r_{energy} - penalty - 0.1 * |\omega_{yaw}|^2
\label{eq:reward}
\end{align}
\begin{equation}
      cost =
    \begin{cases}
      0.8*\nu_{d}-\nu_{x}+penalty & if~\nu_{x} < 0.8*\nu_{d}\\
      \nu_{x}-1.2*\nu_{d}+penalty & if~\nu_{x} > 1.2*\nu_{d}\\
    \end{cases} 
\label{eq:cost}
\end{equation}
where the $r_{energy} = \sum_{i}{\max{(\mathbf{\tau_{i}} * \dot{q}_{i}, 0)}}$, for $i$ in each joint. $\omega_{yaw}$ is the yaw velocity of the robot.
The reason why we purge all negative energy terms is to prevent the agent from exploiting the policy that is caused by inaccurate simulation, which will allow the robot to "generate" power.
The penalty term of the reward function and the cost function is set to 800 and 10 respectively, and is added when the robot looses balance.
\paragraph{Results}
\begin{figure}[ht]
    \centering
    \smallskip
    \includegraphics[width=\columnwidth]{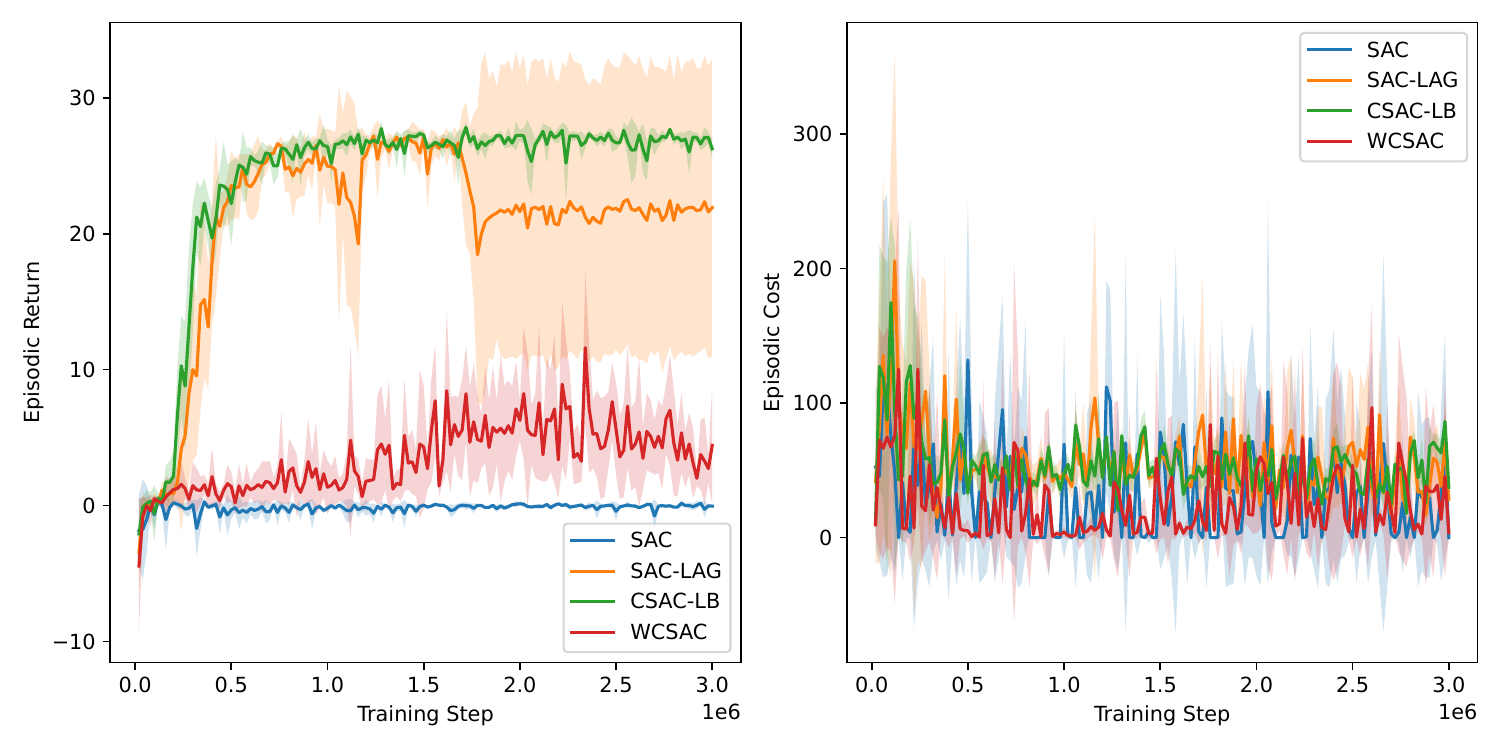}
    \caption{The mean and standard deviation of the episodic return and cost in PointGoal1-v0 environment. Each baseline is trained with 5 seeds for 3e6 environmental steps and is evaluated every 2e4 training steps. CSAC-LB is able to achieve the best performance without a degrading of performance over time as seen for SAC-Lag.}
    \label{fig:res_pointgoal}
\end{figure}
\begin{figure}[ht!]
    \centering
    \begin{subfigure}[b]{0.48\columnwidth}
        \centering
        \includegraphics[width=\textwidth]{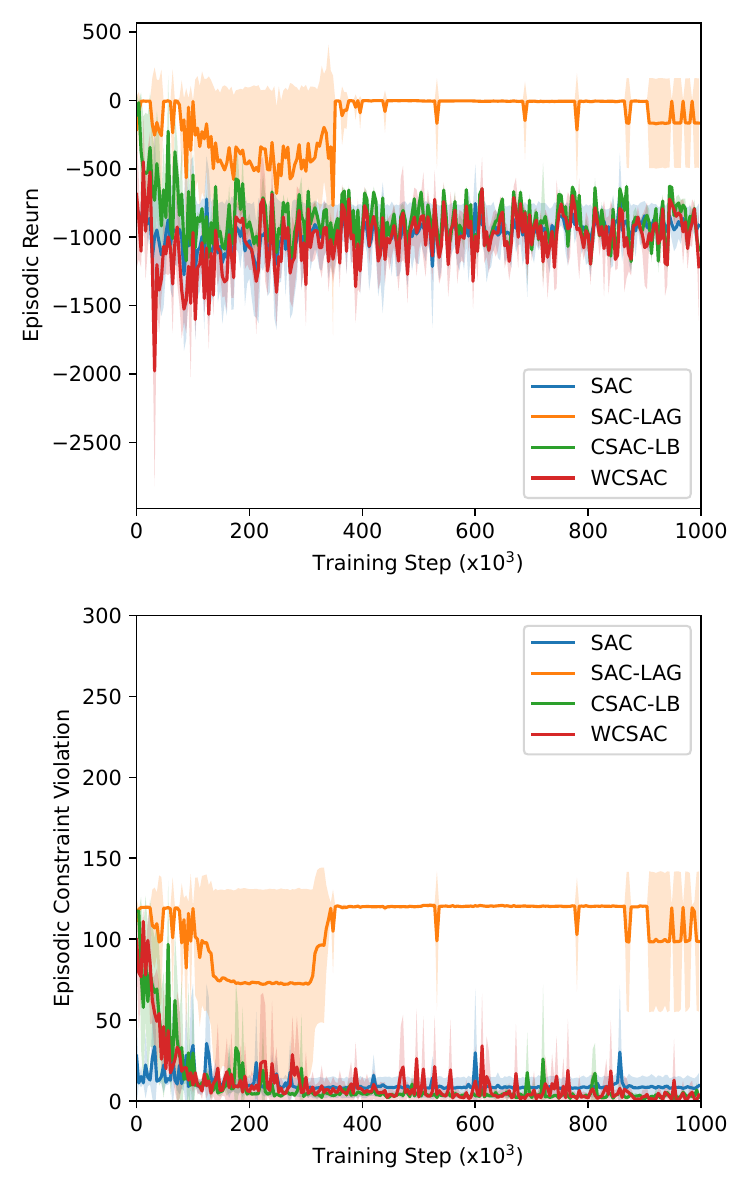}
        \caption{Locomotion, $0.375m/s$}
        \label{fig:slow_speed}
    \end{subfigure}
    \hfill
    \begin{subfigure}[b]{0.48\columnwidth}
        \centering
        \includegraphics[width=\textwidth]{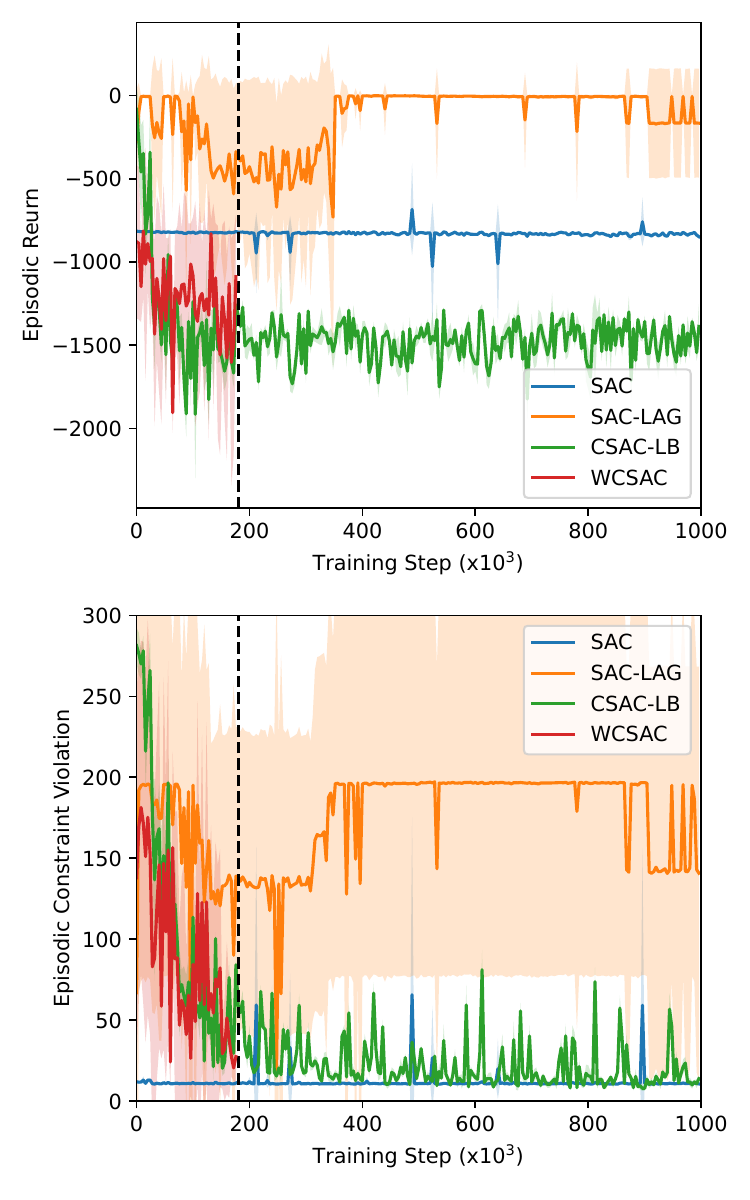}
        \caption{Locomotion, $0.9m/s$}
        \label{fig:high_speed}
    \end{subfigure}
    \caption{The mean and standard deviation of the episodic return (\textbf{Upper}) and
cost (\textbf{Lower}) in locomotion tasks. All agents are trained with 5 seeds for 1e6 training steps and evaluated every 4e3 steps. Only CSAC-LB (green) learns to solve both tasks. WCSAC stops training at the dashed line in right figures due to numerical issues during the training. SAC only learns to lose balance and terminate the episode, constantly violating the constraints.}
    \label{fig:locomotion_results}
\end{figure}
\begin{figure}[h]
    \centering
    \vspace{6pt}
    \begin{subfigure}[b]{\columnwidth}
        \centering
        \includegraphics[width=\textwidth]{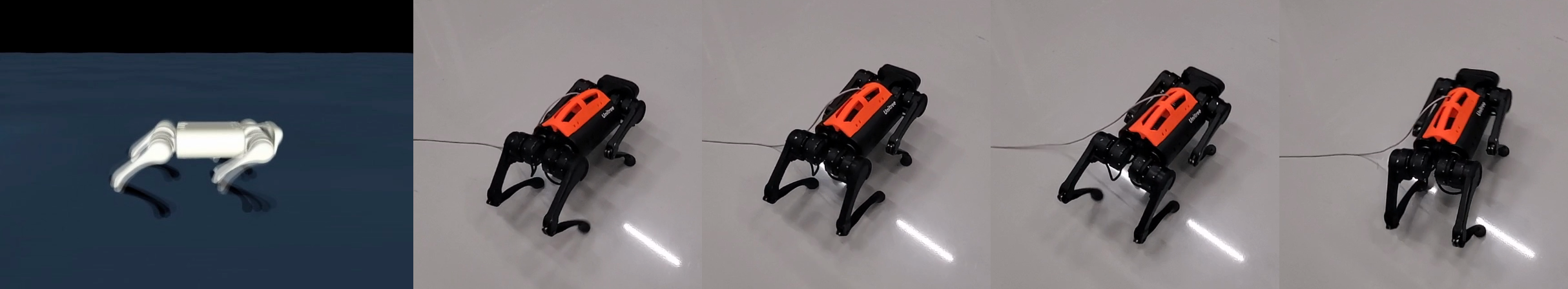}
        \caption{Walking gait learned from low speed constraint}
        \label{fig:walk_new}
    \end{subfigure}
    \hfill
    \begin{subfigure}[b]{\columnwidth}
        \centering
        \includegraphics[width=\textwidth]{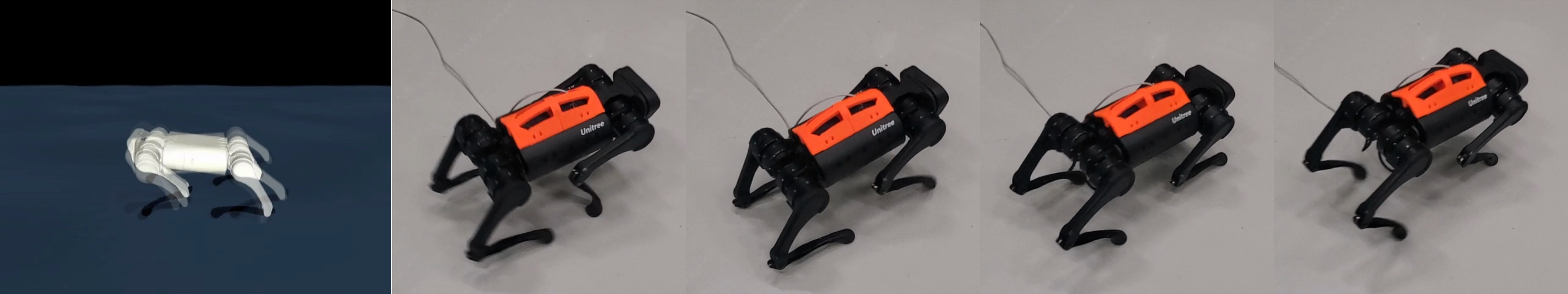}
        \caption{Galloping gait learned from high speed constraint}
        \label{fig:gallop_new}
    \end{subfigure}
    \caption{Two gaits learned by CSAC-LB accordingly when different speeds are specified as constraints. When low speed is desired, the gait is close to walking as each of its leg leaves sequentially. When a higher speed is desired, the robot starts to use both of its feet to accelerate.}
    \label{fig:gait_new}
    \vspace{-5pt}
\end{figure}
Results of PointGoal1-v0 environment are shown in Fig.~\ref{fig:res_pointgoal}. Although SAC and WCSAC-0.5 both have fewer constraint violations compared to SAC-Lag and CSAC-LB, they fail to learn any performing policy and remaining in the initial position. SAC-Lag is able to learn a good policy, just as CSAC-LB. However, SAC-Lag is not able to train robustly and its performance degrades in the later stages of training. This is because in this environment, transitions without constraint violations are much more common than transitions with constraint violations. Therefore, the Lagrange multiplier is updated to a small number (close to 0 in many runs), which leads to unstable training with a risk of ignoring the constraints during the training. In contrast, CSAC-LB makes use of the log barrier function which lets it effectively explore the safe boundary and thus avoids the problem.

For the locomotion tasks, we first plot training results in simulation in Figure~\ref{fig:locomotion_results}. CSAC-LB, SAC and WCSAC manage to learn a reasonable policy when the low-speed constraint is added. The latter two, however, both fail in the high-speed case. The SAC agent stays in a local minimum and learns only to accelerate immediately, then loses balance and terminates the episode early. WCSAC has numerical issues during backpropagation and some of the runs stop at $\sim 20\%$ of the total training steps. SAC-Lag only learns a conservative policy in both cases, in which the agent learns to keep the starting pose and does not take any further action to avoid a penalty from early termination. Only CSAC-LB succeeds in reliably learning policies that satisfy constraints for different constraint values.

Interestingly, as shown in Fig.~\ref{fig:gait_new}, based on the different speed constraints, the CSAC-LB agent actually learns different gaits, namely walking and galloping in different speed constraints (cf.~\cite{fuminimizing} and findings in the field of Biomechanics~\cite{horse}).

\begin{figure}[ht]
\vspace{4pt}
    \centering
    \includegraphics[width=\columnwidth]{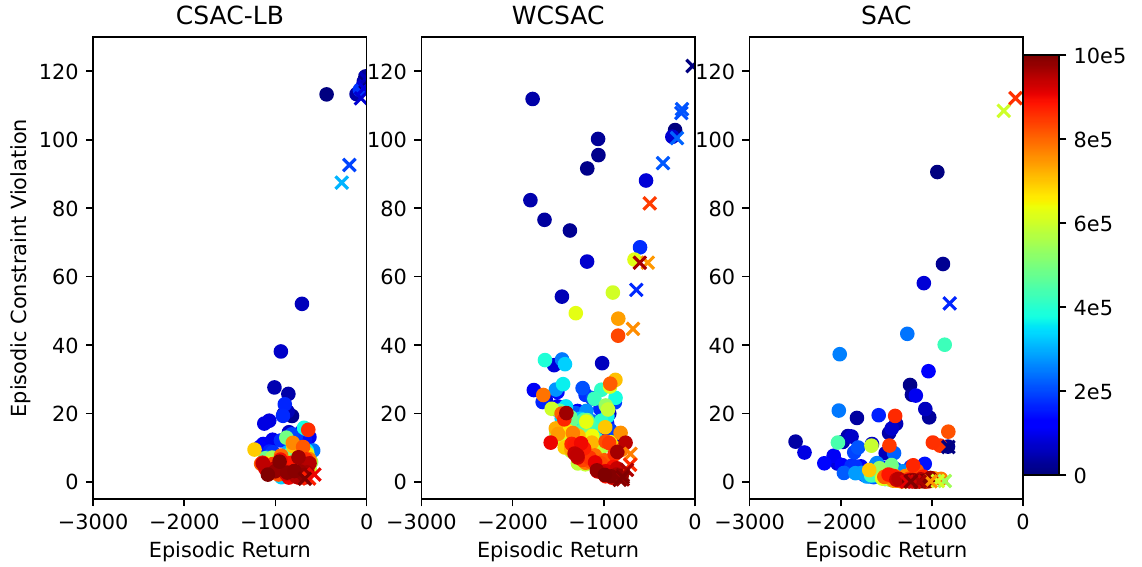}
    \caption{Illustration of the evaluation results during training with $0.375m/s$ as speed constraint. Each data point represents one evaluation episode and its color represents the number of training steps. The $y$-axis is the constraint violation cost and the $x$-axis is the episodic return. Crosses mark points from the Pareto front. SAC-Lag results omitted as no meaningful policy was learned.}
    \label{fig:pareto}
    \vspace{-4pt}
\end{figure}

Fig.~\ref{fig:pareto} indicates that CSAC-LB converges to the best Pareto efficient solution (with the greatest return and lowest constraint violations) compared with the other baselines. Besides, CSAC-LB possesses a more stable training process while the other two still have zig-zagging behaviours that lead to an evaluation further away from the Pareto line. Furthermore, CSAC-LB stays along the safe margin area for longer (close to zero constraint violations), which explains the efficiency of CSAC-LB by learning in more significant regions.


\subsection{Real-robot Experiments}
To test robustness, we also test the learned policies of different algorithms on a Unitree A1 robot via a zero-shot sim-to-real transfer, i.e. by directly transferring policies from simulator to the real robots without any fine-tuning.

Our results show that all baseline algorithms fail to finish the task. Only the policy learned by CSAC-LB can perform normal locomotion skills after training in simulation. This highlights the importance of exploring the safety boundary and the robustness of our algorithm. An example video is provided for more details. In the simulated environment, the rubber feet on the real robot are not properly simulated, thus the agent tends to use less force than is required in practice to accelerate. Therefore, in the video, we can see that the agent cannot accelerate much via jumping forward using its two legs as learned in simulations.

%% file: 7_conclusion.tex
In this paper, we introduce a new Deep RL algorithm suitable for constrained optimization settings. We achieve this by optimizing a linear smoothed log barrier function to better handle numerical issues, and applying it to SAC with Safety Critic, resulting in the CSAC-LB algorithm . We also provide a theoretical performance guarantee bound.
Our method is not limited to combination with SAC, but could also be applied to other RL algorithms such as DDPG, or TD3.

Our evaluation shows the general-purpose applicability of CSAC-LB, while not requiring pre-training, learning of a dynamics model, reward shaping, offline data collection, or extensive hyperparameter tuning.
Our experiments demonstrate its potential in high-dimensional tasks without suffering from degrading of the policy caused by unstable training, which affects Lagrange multiplier methods such as SAC-Lag. A challenging real-world locomotion task also shows the robustness of our algorithm and the importance of effectively exploring the safe margin for learning a safe policy with high returns.

In the future, we aim to develop a mechanism to adaptively adjust the log barrier factor, to further improve the data efficiency of our method. Also, applying the algorithms to other safety-critical domains would be valuable. 